\documentclass{article}

\usepackage[preprint]{neurips_2026}

\usepackage[utf8]{inputenc}
\usepackage[T1]{fontenc}
\usepackage{hyperref}
\usepackage{url}
\usepackage{booktabs}
\usepackage{amsfonts}
\usepackage{amssymb}
\usepackage{amsmath}
\usepackage{nicefrac}
\usepackage{microtype}
\usepackage{graphicx}
\usepackage{xcolor}
\usepackage{placeins}
\usepackage{tikz}
\usetikzlibrary{arrows.meta,positioning,fit,calc}

\graphicspath{{figures/}}
\newcommand{\hitat}[1]{hit@#1}
\newcommand{\mix}[1]{\texttt{#1}}

\workshoptitle{2nd Workshop on Grounded and Faithful Vision-Language Models for Real-World Deployment (VLM4RWD)}

\title{Decoupling What from Where: How Should a Small GUI Grounding Model Receive the Action Type?}

\author{%
  Aadi Chauhan\thanks{Equal contribution.} \\
  Stanford University \\
  \texttt{aadic@stanford.edu} \\
  \And
  Arthur Ilyasov\footnotemark[1] \\
  Stanford University \\
  \texttt{ailyasov@stanford.edu} \\
}

\begin{document}

\maketitle

\begin{abstract}
A GUI agent decides which action to take and where on the screen to take it; we ask how a small grounding model should receive the action type, and what the type buys it. Fine-tuning Qwen2-VL-2B with LoRA on Android in the Wild, we compare a flat baseline with five ways of supplying the type under matched data, compute, adapters, and decoding: an auxiliary loss, a hard-routed action word, an additive learned embedding, a prepended learned token, and the type written into the prompt. With five seeds, a bootstrap that clusters validation steps by episode, and seed-level paired tests, the ranking on a stream of clicks, scrolls, and type events is clear: the auxiliary loss, the additive embedding, and the prompt word each gain five to seven \hitat{0.10} points over the baseline, and hard routing and the prepended token are not distinguishable from it. Much of that gain is protection from a preprocessing choice of ours rather than a spatial prior. Our serializer clamps the off-screen touch point that AITW records for type events to the origin; that class degrades the flat baseline's click grounding, and retraining without it lifts the baseline by nearly seven points, after which no mechanism's hit rate beats it and the intervals are tight enough to exclude a two-point effect, though the auxiliary loss still shortens the average miss; on a stream of taps and swipes none helps. Whether this is a quirk of one serialization or a general hazard is open. For deployment, the pipeline's margin over the baseline with predicted rather than gold types is not established ($+0.016$, 95\% interval $[-0.017, +0.052]$), and a wrong type collapses every model conditioned at inference. The prepended token does not help at the shared learning rate, where its rows barely move from initialization; trained ten times faster it reaches the level of the other three, with a margin over the baseline that three seeds do not establish. We also document a silent failure: conditioning injected through \texttt{inputs\_embeds} makes Qwen2-VL fall back to one-dimensional positions for its image tokens, which cost nine points.
\end{abstract}

\section{Introduction}
\label{sec:intro}

A GUI agent maps a screenshot and an instruction to a low-level action: an action type such as \emph{click}, \emph{scroll}, or \emph{type}, and, for spatial actions, a target location. Native agent models (UI-TARS~\citep{uitars}, OS-Atlas~\citep{osatlas}, SeeClick~\citep{seeclick}, ShowUI~\citep{showui}) produce both in a single autoregressive stream. The two decisions are different kinds of prediction: the type is a small classification problem, and the location is a localization problem that has to find the evidence on the screen and commit to a point. Several agents therefore predict the type first and the target conditioned on it~\citep{cocoagent,limac,hiagent}, and CoCo-Agent ablates its conditional prediction. What is missing is a comparison of the ways of delivering the type under matched data and compute, and a test of whether the trained model reads the signal at all.

This paper is that comparison, in a small model. We fine-tune Qwen2-VL-2B~\citep{qwen2vl} with LoRA~\citep{lora} to emit a coordinate string and compare a flat baseline with five conditioning mechanisms that see the same data in the same order under the same optimizer, adapters, and decoding. We started from the hypothesis that the prepended token would be the right mechanism, because it gives the model a dedicated, attendable position for the type.

The ranking is clear and our hypothesis was wrong. On a stream that mixes clicks, scrolls, and type events, the auxiliary loss, the additive embedding, and the prompt word raise \hitat{0.10} by five to seven points over the baseline (five seeds, episode-clustered intervals, seed-level $p \le 0.03$), almost entirely on clicks; hard routing and the prepended token are within noise of it. The gain comes from a preprocessing choice of ours: AITW records type events with an off-screen touch point that our serializer clamps to the origin, so the flat model must predict a fixed point for a class it is not told about. Exact emissions of that point explain little of the gap; the class pulls the baseline's clicks toward the origin more broadly, retraining without it raises the baseline by nearly seven points, after which no mechanism beats it on the headline metric by more than 0.002, and on a separate stream of taps and swipes none helps. With a Stage-1 classifier at 84\% accuracy the pipeline's margin over the flat baseline is not established, and a wrong type collapses every model conditioned at inference.

Interventions on the trained models separate training effects from inference effects: a wrong or blended type costs both embedding models most of their accuracy, so both read their embedding, and the zeroed additive model is no more than a point below the flat baseline whereas the zeroed prepended-token model is seven points below it. The prepended token's learned rows barely move from their random initialization under the shared learning rate; trained ten times faster they move, and the token reaches the level of the other three mechanisms, so its failure is a statement about the schedule, not the architecture. An earlier implementation that injected the conditioning through \texttt{inputs\_embeds} lost nine points to a silent positional-encoding fallback that we first read as evidence against conditioning.

Our contributions are (1) a matched-compute comparison of five ways of delivering the action type, with five seeds, paired statistics at the episode and seed levels, and a control stream; (2) a diagnosis of where the mixed-stream gain comes from, including retraining with the degenerate class removed; (3) a wrong, zero, and class-mean intervention protocol that tests whether a conditioning signal is read and what remains when it is removed; and (4) an account, checked against the library source, of a positional-encoding failure that inverted a conclusion.

\section{Related Work}
\label{sec:related}

\paragraph{GUI grounding and agent models.}
Mind2Web~\citep{mind2web} and Android in the Wild~\citep{aitw} supply our episodes; AndroidControl~\citep{androidcontrol} and AndroidWorld~\citep{androidworld} extend them. SeeClick~\citep{seeclick} introduced ScreenSpot (extended by ScreenSpot-Pro~\citep{screenspotpro}), and UGround~\citep{uground} and OS-Atlas~\citep{osatlas} scale grounding pretraining. Aguvis~\citep{aguvis}, UI-TARS~\citep{uitars}, CogAgent~\citep{cogagent}, ScreenAI~\citep{screenai}, and Ferret-UI~\citep{ferretui} emit action and location together, the design our flat baseline follows; TinyClick~\citep{tinyclick} and Ferret-UI Lite~\citep{ferretuilite} work at our scale. GUI-Actor~\citep{guiactor} replaces coordinate text with an attention head, MolmoPoint~\citep{molmopoint} and GUI-AIMA~\citep{guiaima} add grounding or anchor tokens, UI-R1~\citep{uir1} rewards type and argument separately, and OmniParser~\citep{omniparser} and Set-of-Mark~\citep{som} externalize grounding. \citet{imrope} argue that the implicit mapping from rotary position to coordinate limits grounding and add explicit coordinate markers with an interleaved M-RoPE; Section~\ref{sec:mrope} shows a cruder way the same positions go wrong.

\paragraph{Type-first agents.}
CoCo-Agent~\citep{cocoagent} predicts the type and then the target conditioned on it in one stream, our hard-routing variant. LiMAC~\citep{limac} places a small action transformer ahead of a LoRA-tuned Qwen2-VL-2B on AITW and AndroidControl; the action transformer predicts the type and selects click targets contrastively over UI-element embeddings, and the VLM generates only text arguments. AutoGLM~\citep{autoglm} and Hi-Agent~\citep{hiagent} separate planning from low-level action, as do SeeAct~\citep{seeact}, Ponder\,\&\,Press~\citep{ponderpress}, and Aria-UI~\citep{ariaui}. None holds data, compute, and decoding fixed across ways of delivering the type, or intervenes on the trained model to ask whether the signal is used.

\paragraph{Conditioning, faithfulness, and the backbone.}
SayCan~\citep{saycan}, ReAct~\citep{react}, and RT-2~\citep{rt2} factor, expose, or tokenize the action decision; AutoUI~\citep{autoui}, AppAgent~\citep{appagent}, DigiRL~\citep{digirl}, WebRL~\citep{webrl}, VSC-RL~\citep{vscrl}, and GUI-Libra~\citep{guilibra} improve mobile agents without changing how the type enters grounding, and our auxiliary loss is multi-task learning in the sense of \citet{caruana}. Kosmos-2~\citep{kosmos2}, Shikra~\citep{shikra}, and Ferret~\citep{ferret} teach VLMs to refer to regions; POPE~\citep{pope} and HallusionBench~\citep{hallusionbench} ask whether an output is supported by the image and our attention measurements ask that of a coordinate, while \citet{prefill} report that grounding is largely fixed during prefill. Qwen2-VL~\citep{qwen2vl} assigns three-dimensional rotary positions (extending RoPE~\citep{roformer}) to visual tokens from \texttt{input\_ids}; Section~\ref{sec:mrope} describes what happens when conditioning bypasses that path.

\section{Method}
\label{sec:method}

\begin{figure}[t]
  \centering
  \resizebox{0.86\linewidth}{!}{
\begin{tikzpicture}[
  font=\footnotesize,
  >={Stealth[length=1.8mm]},
  box/.style={draw, rounded corners=2pt, align=center, inner sep=2.5pt, minimum height=7mm},
  inp/.style={box, fill=gray!12, text width=1.55cm},
  model/.style={box, fill=blue!10, thick, text width=2.25cm},
  head/.style={box, fill=orange!18, text width=1.35cm},
  outp/.style={box, fill=green!14, text width=1.6cm},
  mech/.style={box, fill=purple!8, text width=1.95cm, minimum height=11mm, font=\scriptsize},
  lbl/.style={font=\scriptsize\itshape, text=black!65},
  cond/.style={->, thick, purple!70!black},
  node distance=3mm,
]
  \node[inp] (shot) {screenshot\\ + goal};
  \node[model, right=4mm of shot] (vlm1) {Qwen2-VL-2B\\ \textit{frozen}, mean-pooled};
  \node[head, right=of vlm1] (mlp) {MLP\\ (3 layers)};
  \node[outp, right=of mlp] (atype) {action type\\ $\hat a$};
  \node[model, right=10mm of atype] (vlm2) {Qwen2-VL-2B\\ + LoRA on q,k,v,o};
  \node[outp, right=of vlm2] (coord) {\texttt{(x, y)}\\ 0--999 grid};
  \node[draw, densely dotted, rounded corners, inner sep=3pt, fit=(vlm1)(mlp)(atype),
        label={[lbl]below:{Stage 1: \textbf{what} (action type)}}] (s1) {};
  \node[draw, densely dotted, rounded corners, inner sep=3pt, fit=(vlm2)(coord),
        label={[lbl]below:{Stage 2: \textbf{where} (grounding)}}] (s2) {};
  \draw[->] (shot) -- (vlm1);
  \draw[->] (vlm1) -- (mlp);
  \draw[->] (mlp) -- (atype);
  \draw[->] (vlm2) -- (coord);
  \draw[cond] (atype) -- node[above, font=\scriptsize, text=purple!70!black, align=center]
    {gold type, or\\ $\hat a$ at test} (vlm2);
  \draw[->] (shot.north) -- ++(0,6mm) -| (vlm2.north);
  \node[lbl, anchor=south] at ($(shot.north)!0.5!(vlm2.north) + (0,6mm)$) {same screenshot + goal};

  \coordinate (mid) at ($(shot.west)!0.5!(coord.east)$);
  \node[mech, anchor=north east] (C) at ($(mid) + (-1mm,-13mm)$)
    {\textbf{C} hard routing\\ emit \texttt{click (x, y)};\\ word forced at test};
  \node[mech, anchor=north west] (Dh) at ($(mid) + (1mm,-13mm)$)
    {\textbf{D-hook} additive\\ $E[a]$ added to text\\ embeddings; zero init};
  \node[mech, left=2mm of C] (B)
    {\textbf{B} aux.\ loss\\ linear head on pooled\\ state; training only};
  \node[mech, left=2mm of B] (A)
    {\textbf{A} flat\\ no type\\ (baseline)};
  \node[mech, right=2mm of Dh] (Dt)
    {\textbf{D-token} prepended\\ \texttt{<|action\_slot|>}\\ $\to E[a]$; random init};
  \node[mech, right=2mm of Dt] (Dx)
    {\textbf{D-text} prompt word\\ \texttt{Action: click.}\\ no new parameters};
  \node[draw, densely dotted, rounded corners, inner sep=3pt, fit=(A)(B)(C)(Dh)(Dt)(Dx),
        label={[lbl]below:{how the type enters Stage 2 (same data, optimizer, adapters, decoding)}}] (fitbox) {};
  \draw[cond] (fitbox.north -| vlm2.south) -- (vlm2.south);
\end{tikzpicture}}
  \caption{Two-stage pipeline. Stage~1 classifies the action type from mean-pooled frozen features. Stage~2 grounds the action to a coordinate string, and the action type enters it through one of the six mechanisms in the bottom row, all of which share data, optimizer, adapters, and decoding. D-slot, the retracted variant whose \texttt{inputs\_embeds} injection bypassed the position computation, is described in Section~\ref{sec:mrope}.}
  \label{fig:arch}
\end{figure}
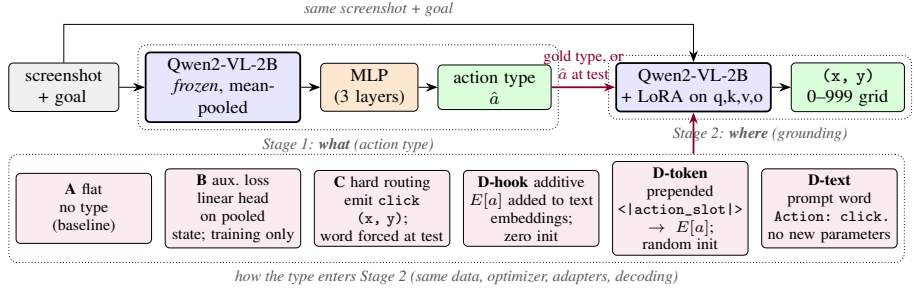

\subsection{Action taxonomy and grounding target}
Both datasets are mapped onto a canonical eight-class action space (\emph{click}, \emph{double-click}, \emph{type}, \emph{scroll}, \emph{drag}, \emph{hotkey}, \emph{wait}, \emph{finished}). Mind2Web's CLICK, SELECT, and HOVER map to \emph{click} and TYPE to \emph{type}. AITW stores gestures as touch and lift points; we split them into taps (\emph{click}) and swipes (\emph{scroll}) by the normalized distance between the two, with threshold 0.04. The grounding target is the touch point normalized to $[0,1]^2$, serialized as \texttt{(x, y)} with integers on a $0$--$999$ grid so every target has the same token length. AITW records non-touch actions such as typing with the touch point $(-1,-1)$; the serializer clamps this to \texttt{(0, 0)}, so every \emph{type} event in the training stream has the same target at the screen origin. Its distance from the raw target is at least $\sqrt{2}$ for any on-screen prediction, so no model can score a hit on a type event; we keep these events in the validation slices, where every model scores zero on them, and Section~\ref{sec:diagnosis} measures what they do to training.

\subsection{Stage 1: which action}
Stage~1 runs the frozen Qwen2-VL-2B-Instruct backbone over the screenshot and instruction, mean-pools the final hidden states over unmasked tokens into a 1536-dimensional vector, and trains a three-layer MLP ($1536 \to 1024 \to 256 \to 8$, dropout 0.1) with class-weighted cross-entropy (AdamW, learning rate $10^{-4}$, weight decay 0.01, batch size 128, eight epochs, best validation macro-F1 checkpoint). Features come from text only, text plus screenshot, or a zeroed sanity variant.

\subsection{Stage 2: where}
Stage~2 fine-tunes the same backbone with LoRA adapters (rank 16, $\alpha=32$, dropout 0.05) on the query, key, value, and output projections, 4.4M trainable parameters out of 2.2B. The prompt is the screenshot, \texttt{Goal: \{goal\}}, and a fixed instruction to predict the action coordinate; the answer is the coordinate string. With $s$ the screenshot, $g$ the goal, $a$ the action type, and $y_{1:T}$ the coordinate tokens, every variant minimizes $\mathcal{L}_{\mathrm{coord}} = -\sum_{t=1}^{T} \log p_\theta(y_t \mid y_{<t}, s, g, c)$, masked to the answer tokens, with $c$ the variant-specific conditioning.

\textbf{A, flat.} $c=\varnothing$: the joint-decoding design of native agent models, reduced to the grounding sub-problem.

\textbf{B, auxiliary loss.} A linear head $h_\phi$ on the mean-pooled last hidden state $\bar z$ predicts the action type during training, $\mathcal{L}_{\mathrm{B}} = \mathcal{L}_{\mathrm{coord}} + \lambda\,\mathrm{CE}(h_\phi(\bar z), a)$ with $\lambda=1$. The mean runs over every unmasked position of the teacher-forced sequence, answer tokens included, so the head can read the type off the answer (a type event's answer is \texttt{(0, 0)}). Nothing about the type is provided at inference, so any gain is a training effect.

\textbf{C, hard routing.} The model is trained to emit the action word before the coordinate, \texttt{click (x, y)}, as in CoCo-Agent's conditional action prediction. At test time the word for the gold (or Stage-1) type is forced as a decoding prefix.

\textbf{D-hook, additive embedding.} A learned table $E \in \mathbb{R}^{8\times d}$ holds one vector per class. A forward hook on the text-embedding module adds $E[a]$ to the embedding of every position in \texttt{input\_ids}; Qwen2-VL then overwrites the image-placeholder positions with the visual features, so the bias reaches the text tokens (the prompt and, in training, the answer) and not the image tokens. $E$ is zero-initialized, so D-hook starts as an exact copy of A.

\textbf{D-token, prepended action token.} A new token \texttt{<|action\_slot|>} is added to the vocabulary and placed at the start of the text prompt, so it occupies a real position in \texttt{input\_ids}; a forward hook replaces its embedding with $E[a]$, initialized from $\mathcal{N}(0, 0.02^2)$. This is the architecture we originally hypothesized: a full-norm, attendable representation of the type. The table shares the base learning rate; Section~\ref{sec:headline} also trains it ten and a hundred times faster.

\textbf{D-text, type as a word.} The prompt begins with \texttt{Action: click.} (or the gold type's word), using the pretrained word embeddings and no new parameters: the obvious baseline for the two learned-embedding variants.

For C, D-hook, D-token, and D-text the gold type is used in training and at evaluation unless stated otherwise; Section~\ref{sec:e2e} closes the loop with Stage-1 predictions. All variants share the data order, the optimizer (AdamW, learning rate $2\times10^{-5}$, weight decay 0.01, batch size 1, gradient clipping at 1.0, two epochs), and greedy decoding with 16 new tokens. No hyperparameter was tuned per variant.

\section{Experimental setup}
\label{sec:setup}

\paragraph{Data.}
Stage~1 uses 7,775 Multimodal-Mind2Web training steps evaluated on the 1,338-step \texttt{test\_task} split, and 5,000 AITW General steps evaluated on 1,000 held-out steps. Stage~2 uses two AITW streams, built by filtering the training split of the \texttt{cjfcsjt/AITW\_General} mirror in its stored order. \mix{all\_with\_coords} keeps taps, swipes, and type events; at 1,200 training examples its validation slice of 250 steps holds 169 clicks, 46 scrolls, and 35 type events from 42 episodes. \mix{taps\_and\_swipes} keeps only taps and swipes (1,000 training and 200 validation steps from 37 episodes), which land anywhere on the screen. For a given mix and size every seed and variant sees the same examples in the same order, which makes paired tests possible; the validation slice begins where the training slice ends, so absolute numbers are not comparable across sizes. This is not the official episode split and no action history is given, which limits comparability with published numbers, not the matched comparison. A second control uses Multimodal-Mind2Web (1,000 training and 250 evaluation steps; the target is the gold element's center). Stage~2 is studied on AITW because there the type is a vision problem: Stage~1 gains $+0.414$ macro-F1 from the screenshot on AITW and $+0.009$ on Mind2Web, whose instructions name the action (Table~\ref{tab:stage1}).

\paragraph{Metrics.}
\hitat{r} is the fraction of predictions within normalized Euclidean distance $r$ of the target, for $r \in \{0.05, 0.10, 0.25\}$ and $r = 0.14$, the tap threshold of AITW's matcher~\citep{aitw}; \hitat{0.10} is the headline, and we report the mean normalized distance. Unparseable outputs count as misses at distance $\sqrt{2}$; only hard routing produces them, and Section~\ref{sec:headline} reports its numbers on parsed outputs as well.

\paragraph{Statistics.}
The headline setting uses five seeds (42 to 46); every other cell uses three (42 to 44), and we report the mean and sample standard deviation over seeds. Every run logs a per-example distance on a shared evaluation set with a shared per-seed data order, so comparisons are paired at two levels. Steps of one episode share app, goal, and near-identical screens, so our primary interval is a cluster bootstrap over episodes, resampling episodes with replacement and keeping every step and seed of a sampled episode together (10,000 resamples, 95\% percentile interval, two-sided bootstrap $p$~\citep{efron}). One episode straddles the training/validation boundary in every \mix{all\_with\_coords} slice; dropping its steps moves no estimate by more than 0.003 (Appendix~\ref{app:details}). A paired $t$-test on the five (or three) per-seed means has little power but cannot be fooled by within-seed correlation; we call a difference established only when both agree, and stars come from the episode bootstrap with the seed-level $p$ beside them. No correction is made for the more than fifty contrasts reported; the scaling study (Appendix~\ref{app:scaling}) is descriptive for that reason.

\section{Results}
\label{sec:results}

\subsection{On the mixed stream, three mechanisms beat the flat baseline and two do not}
\label{sec:headline}

\begin{table}[t]
\centering
\small
\caption{Stage 2 grounding on AITW \texttt{all\_with\_coords} (n$_{\text{train}}$=1200, n$_{\text{val}}$=250, 5 seeds). Mean $\pm$ std over seeds; $\Delta$ vs.\ A with a 95\% episode-cluster bootstrap CI (stars from the same bootstrap, $^{*}p<.05$, $^{**}p<.01$, $^{***}p<.001$) and the seed-level paired $t$-test $p$.}
\label{tab:headline}
\resizebox{\linewidth}{!}{\begin{tabular}{lcccccc}
\toprule
Variant & hit@0.05 & hit@0.10 & hit@0.25 & mean L2 $\downarrow$ & $\Delta$hit@0.10 vs.\ A [episode CI] & seed-level $p$ \\
\midrule
A: flat & 0.130 $\pm$ 0.039 & 0.229 $\pm$ 0.043 & 0.499 $\pm$ 0.029 & 0.406 $\pm$ 0.028 &  &  \\
B: aux.\ loss & 0.178 $\pm$ 0.025 & 0.293 $\pm$ 0.021 & 0.582 $\pm$ 0.026 & 0.365 $\pm$ 0.004 & +0.064 [+0.037, +0.092]$^{***}$ & 0.004 \\
C: hard routing & 0.170 $\pm$ 0.022 & 0.264 $\pm$ 0.045 & 0.538 $\pm$ 0.059 & 0.421 $\pm$ 0.016 & +0.035 [-0.002, +0.072] & 0.189 \\
D-hook: additive & 0.170 $\pm$ 0.042 & 0.282 $\pm$ 0.041 & 0.563 $\pm$ 0.016 & 0.370 $\pm$ 0.010 & +0.054 [+0.022, +0.090]$^{***}$ & 0.028 \\
D-token: prepended & 0.144 $\pm$ 0.056 & 0.238 $\pm$ 0.049 & 0.504 $\pm$ 0.055 & 0.386 $\pm$ 0.021 & +0.009 [-0.019, +0.036] & 0.465 \\
D-text: word in prompt & 0.186 $\pm$ 0.022 & 0.302 $\pm$ 0.028 & 0.568 $\pm$ 0.033 & 0.364 $\pm$ 0.008 & +0.073 [+0.030, +0.117]$^{**}$ & 0.018 \\
\bottomrule
\end{tabular}}
\end{table}

Table~\ref{tab:headline} is the main comparison: six variants on \mix{all\_with\_coords} at 1,200 training examples, five seeds each, with episode-clustered intervals. Three mechanisms beat the flat baseline on every hit rate and on mean distance: the auxiliary loss (B) by $+0.064$ \hitat{0.10} (95\% interval $[+0.037, +0.092]$, seed-level $p = 0.004$), the additive embedding (D-hook) by $+0.054$ ($[+0.022, +0.090]$, $p = 0.028$), and the type written into the prompt (D-text) by $+0.073$ ($[+0.030, +0.117]$, $p = 0.018$). The three are indistinguishable (D-hook $-$ B: $-0.010$, $[-0.043, +0.024]$; D-text $-$ B: $+0.009$, $[-0.040, +0.058]$). Hard routing (C) gains $+0.035$ with an interval through zero ($[-0.002, +0.072]$, $p = 0.19$) and loses on mean distance because some forced outputs do not parse (Table~\ref{tab:perclass}). The prepended token (D-token), the mechanism we set out to validate, is within noise of the baseline on every hit rate ($+0.009$, $[-0.019, +0.036]$) and below B by $-0.055$ ($[-0.083, -0.026]$, $p = 0.016$). At AITW's 0.14 tap threshold the picture is the same (B $+0.059$, D-hook $+0.042$, D-text $+0.067$, intervals excluding zero and seed-level $p \le 0.04$; C $+0.018$ and D-token $+0.003$ with intervals through zero). Seed variance is large (the baseline spans 0.169 to 0.280), hence paired statistics rather than a best run. The headline cell had three seeds when the study was designed; the fourth and fifth seeds and D-text were added after the three-seed ordering was known, and it did not change.

D-token's failure is confounded by an untrained embedding: at the shared learning rate its rows move by 0.016 from an initialization of norm 0.78 (Table~\ref{tab:dtoken_lr}), so Table~\ref{tab:headline} tests a fixed random class code plus LoRA. Training the table ten and a hundred times faster moves the rows by 0.28 and 1.9 and raises \hitat{0.10} to 0.295 and 0.283 (three seeds), within noise of B; the margin over the baseline, $+0.040$ ($[+0.005, +0.076]$, seed $p = 0.16$) and $+0.028$ ($[-0.008, +0.065]$, seed $p = 0.02$), passes one of our two tests each time and not the other. We therefore call the trained token level with the other conditioned models and its advantage over the baseline suggestive: the refutation is of the schedule, not the architecture.

\begin{table}[t]
\centering
\small
\caption{Where the gain lives. Per-class \hitat{0.10} on the headline mix (five seeds, same basis as Table~\ref{tab:headline}); hard routing's unparsed share and its rates on parsed outputs; the fraction of click predictions that are the clamped type target \texttt{(0, 0)}, and click \hitat{0.10} excluding them. Every model scores zero on the 35 \emph{type} events because the raw target is off-screen.}
\label{tab:perclass}
\resizebox{\linewidth}{!}{\begin{tabular}{lcccccc}
\toprule
Variant & click & scroll & unparsed click / scroll & click, parsed only & clicks at \texttt{(0, 0)} & click, excluding those \\
\midrule
A: flat & $0.256 \pm 0.062$ & $0.304 \pm 0.034$ & 0\% / 0\% & $0.256$ & $4.3\%$ (13\% on the worst seed) & $0.266$ \\
B: aux.\ loss & $0.357 \pm 0.026$ & $0.278 \pm 0.018$ & 0\% / 0\% & $0.357$ & $2.4\%$ & $0.366$ \\
C: hard routing & $0.350 \pm 0.061$ & $0.148 \pm 0.073$ & 1.4\% / 13.9\% & $0.356$ (scroll $0.172$) & $0.0\%$ & $0.350$ \\
D-hook: additive & $0.321 \pm 0.064$ & $0.357 \pm 0.039$ & 0\% / 0\% & $0.321$ & $0.0\%$ & $0.321$ \\
D-token: prepended & $0.269 \pm 0.069$ & $0.304 \pm 0.090$ & 0\% / 0\% & $0.269$ & $0.0\%$ & $0.269$ \\
D-text: prompt word & $0.348 \pm 0.036$ & $0.361 \pm 0.036$ & 0\% / 0\% & $0.348$ & $0.0\%$ & $0.348$ \\
\bottomrule
\end{tabular}}
\end{table}

\subsection{How much of the gain is protection from a degenerate class?}
\label{sec:diagnosis}

Table~\ref{tab:perclass} shows where the gain lives. It is a click gain: B and D-text gain nine to ten points on clicks, D-hook six, and C ten while paying sixteen on scrolls, because forcing the word \emph{scroll} derails the gesture (it fails to parse 14\% of scroll outputs against 1\% of click outputs, and its parsed scroll rate is still 0.172, so most of that deficit is misplacement). D-hook and D-text improve both classes. The gain comes from our own preprocessing: the serializer clamps AITW's $(-1,-1)$ touch point for type events to \texttt{(0, 0)}, so the mixed stream contains a class with a fixed target that the flat model must predict with no indication of the type, a situation no deployed action format creates (native formats emit no coordinate for a type action, or emit the type first). The simplest story, that the baseline emits the clamped target for clicks, accounts for little: it does so for 4.3\% of clicks, B for 2.4\%, and the models conditioned at inference never, and excluding those predictions moves the baseline's click rate only from 0.256 to 0.266. The damage is broader: on the seed for which we logged predictions, the baseline's clicks lie 0.10 of the screen width left of their targets on average (0.15 among misses) and 7\% fall within 0.10 of the origin, against $-0.02$ and 0.6\% for D-hook on the same screens, so the class pulls its clicks toward the clamped point without making it emit that point.

Retraining measures how much of the gain is protection from that class: we retrain all six variants on the same training slice with the type events removed and score the identical 250 validation examples (five seeds for A, B and D-hook, three for the rest; Table~\ref{tab:notype}, appendix). Removing the class raises the flat baseline from 0.229 to 0.296 \hitat{0.10} ($+0.067$, $[+0.044, +0.091]$, seed $p = 0.04$; on clicks from 0.256 to 0.355) and leaves every conditioned variant within 0.025 of where it was, with intervals through zero. On that stream no mechanism beats the baseline on the headline metric: B's margin is $+0.002$ ($[-0.014, +0.018]$, seed $p = 0.93$) and D-hook's $+0.001$ ($[-0.020, +0.022]$, seed $p = 0.97$), intervals tight enough to exclude a two-point effect, while C, D-token and D-text are at $-0.028$, $-0.024$ and $-0.015$. The gain on \hitat{0.10} is therefore accounted for by the class. What survives is smaller and about a different quantity: B shortens the average miss by $-0.023$ ($[-0.030, -0.016]$, seed $p = 0.02$), the only clean-stream contrast that passes both of our tests, and gains $+0.035$ \hitat{0.25} ($[+0.012, +0.059]$, seed $p = 0.16$) on the bootstrap alone. With the control in the next section, we read this as follows: on the mixed stream, protection from a class our serialization made degenerate accounts for the \hitat{0.10} gain, and we do not claim a spatial prior. Two things keep the question open. Type supervision still buys something on a clean stream, just not precise localization: the auxiliary loss reliably shortens the average miss without hitting the target more often, which is what a model with a better prior on plausible regions and no better fix on the exact widget would look like. And the result rests on one dataset, one serialization, and validation slices of a few hundred steps from a few dozen episodes, so we report it as a hazard we found in our own pipeline rather than a general property of type supervision.

\subsection{Control: no gain on taps and swipes}

\begin{table}[t]
\centering
\small
\caption{Control: AITW \texttt{taps\_and\_swipes} (n$_{\text{train}}$=1000, n$_{\text{val}}$=200, 3 seeds), a training stream without the degenerate class. Same statistics as Table~\ref{tab:headline}.}
\label{tab:control}
\resizebox{\linewidth}{!}{\begin{tabular}{lcccccc}
\toprule
Variant & hit@0.05 & hit@0.10 & hit@0.25 & mean L2 $\downarrow$ & $\Delta$hit@0.10 vs.\ A [episode CI] & seed-level $p$ \\
\midrule
A: flat & 0.243 $\pm$ 0.020 & 0.382 $\pm$ 0.015 & 0.720 $\pm$ 0.017 & 0.185 $\pm$ 0.003 &  &  \\
B: aux.\ loss & 0.225 $\pm$ 0.015 & 0.368 $\pm$ 0.012 & 0.730 $\pm$ 0.039 & 0.173 $\pm$ 0.011 & -0.013 [-0.042, +0.013] & 0.456 \\
C: hard routing & 0.217 $\pm$ 0.016 & 0.325 $\pm$ 0.040 & 0.635 $\pm$ 0.064 & 0.271 $\pm$ 0.068 & -0.057 [-0.101, -0.011]$^{*}$ & 0.215 \\
D-hook: additive & 0.235 $\pm$ 0.022 & 0.363 $\pm$ 0.021 & 0.725 $\pm$ 0.043 & 0.182 $\pm$ 0.009 & -0.018 [-0.056, +0.019] & 0.334 \\
D-text: word in prompt & 0.210 $\pm$ 0.022 & 0.335 $\pm$ 0.031 & 0.665 $\pm$ 0.013 & 0.199 $\pm$ 0.003 & -0.047 [-0.094, +0.002] & 0.088 \\
\bottomrule
\end{tabular}}
\end{table}

On \mix{taps\_and\_swipes} (Table~\ref{tab:control}), whose training stream has no degenerate class, B and D-hook are within noise of the baseline ($-0.013$ and $-0.018$ \hitat{0.10}, intervals $[-0.042, +0.013]$ and $[-0.056, +0.019]$), so the data admit harms of up to five points as well as no effect; D-text is worse by $-0.047$, with an interval that also includes zero ($[-0.094, +0.002]$, seed $p = 0.09$). C is worse by $-0.057$ with an interval that excludes zero ($[-0.101, -0.011]$) but a seed-level $p$ of 0.22, and a large increase in mean distance; on parsed outputs its deficit shrinks to $-0.045$ (0.337 against 0.382), so parse failures explain part of it and not all. The Multimodal-Mind2Web control is at floor (a tenth of predicted points inside the gold element for both A and D-hook) and is reported in the appendix only.

\subsection{End to end with predicted types: the deployment number}
\label{sec:e2e}
To close the loop we train D-hook, fit the Stage-1 MLP on frozen features of its training examples (200 full-batch epochs, no checkpoint selection), and evaluate it on the same validation examples with gold and with predicted types (three seeds). Stage~1 reaches 0.843 accuracy against a majority class of 0.676. The oracle pipeline scores 0.292 \hitat{0.10}, the predicted-type pipeline 0.271, the flat baseline 0.255. The predicted pipeline's margin over the baseline is not established ($+0.016$, $[-0.017, +0.052]$, seed $p = 0.44$); the oracle-to-predicted gap is ($+0.021$, $[+0.003, +0.041]$, seed $p = 0.07$). Classifier error removes more than half of the oracle gain, and Section~\ref{sec:interventions} shows why it is the risk: a wrong type costs far more than a missing one.

\section{Mechanism}
\label{sec:mechanism}

\subsection{Interventions: which signals are read, and what remains when they are removed}
\label{sec:interventions}

\begin{table}[t]
\centering
\small
\caption{Interventions: the same trained model decoded with the gold action embedding, a wrong one (cyclic; click$\to$type), a wrong one swapping only clicks and scrolls, the table zeroed, and every row replaced by the class mean. Five seeds $\times$ 250 examples; $\Delta$ = gold $-$ condition, 95\% episode-cluster CI, seed-level paired $p$; flat A on the same examples is the reference row.}
\label{tab:interv}
\resizebox{\linewidth}{!}{\begin{tabular}{llccccc}
\toprule
Model & Conditioning & hit@0.10 & click hit@0.10 & scroll hit@0.10 & $\Delta$ gold $-$ cond.\ (hit@0.10) & seed $p$ \\
\midrule
D-hook: additive & gold type & 0.275 $\pm$ 0.040 & 0.310 $\pm$ 0.060 & 0.357 $\pm$ 0.052 &  &  \\
 & wrong (cyclic) & 0.054 $\pm$ 0.009 & 0.006 $\pm$ 0.000 & 0.270 $\pm$ 0.050 & +0.222 [+0.182, +0.263]$^{***}$ & 0.000 \\
 & wrong (click$\leftrightarrow$scroll) & 0.099 $\pm$ 0.056 & 0.073 $\pm$ 0.087 & 0.270 $\pm$ 0.050 & +0.176 [+0.134, +0.221]$^{***}$ & 0.003 \\
 & zeroed & 0.248 $\pm$ 0.032 & 0.288 $\pm$ 0.053 & 0.291 $\pm$ 0.050 & +0.027 [-0.006, +0.065] & 0.224 \\
 & class mean & 0.206 $\pm$ 0.058 & 0.217 $\pm$ 0.099 & 0.322 $\pm$ 0.064 & +0.070 [+0.034, +0.109]$^{***}$ & 0.133 \\
 & flat A, same examples & 0.229 $\pm$ 0.043 & 0.256 & 0.304 & +0.046 [+0.016, +0.081]$^{**}$ & 0.032 \\
\midrule
D-token: prepended & gold type & 0.242 $\pm$ 0.054 & 0.272 $\pm$ 0.079 & 0.313 $\pm$ 0.099 &  &  \\
 & wrong (cyclic) & 0.079 $\pm$ 0.012 & 0.036 $\pm$ 0.019 & 0.300 $\pm$ 0.054 & +0.162 [+0.127, +0.197]$^{***}$ & 0.001 \\
 & wrong (click$\leftrightarrow$scroll) & 0.155 $\pm$ 0.043 & 0.148 $\pm$ 0.063 & 0.300 $\pm$ 0.054 & +0.086 [+0.050, +0.125]$^{***}$ & 0.073 \\
 & zeroed & 0.162 $\pm$ 0.072 & 0.160 $\pm$ 0.112 & 0.291 $\pm$ 0.033 & +0.080 [+0.054, +0.107]$^{***}$ & 0.120 \\
 & class mean & 0.169 $\pm$ 0.043 & 0.169 $\pm$ 0.067 & 0.296 $\pm$ 0.025 & +0.073 [+0.046, +0.100]$^{***}$ & 0.091 \\
 & flat A, same examples & 0.229 $\pm$ 0.043 & 0.256 & 0.304 & +0.013 [-0.013, +0.039] & 0.374 \\
\bottomrule
\end{tabular}}
\end{table}

D-token might ignore its slot, and D-hook's gain might be a training effect that leaves a vector the model never reads. To test both on equal terms we retrain each at the headline setting (shared learning rate, five seeds) and decode every trained model under five conditionings: the gold type; a wrong type from a cyclic permutation of the classes present (click to type, type to scroll, scroll to click); a second wrong map swapping only clicks and scrolls, so no groundable example is routed to the degenerate class; the table zeroed; and every row replaced by the mean of the rows present. Table~\ref{tab:interv} reports the results with the flat baseline on the same examples (an earlier three-condition test, Table~\ref{tab:causal}, agrees). D-hook's learned rows are small (norm 0.04 against a mean token-embedding norm of 0.58; D-token's have norm 0.78), and the model reads them. A wrong type collapses D-hook from 0.275 to 0.054 under the cyclic map and to 0.099 under the click-scroll map (gold minus wrong: $+0.222$, $[+0.182, +0.263]$, and $+0.176$, $[+0.134, +0.221]$; seed $p < 0.001$ and $0.003$), so the collapse is not an artifact of routing clicks to the degenerate class. The class-mean vector costs seven points ($+0.070$, $[+0.034, +0.109]$), though the seed-level test does not reach significance ($p = 0.13$). Zeroing the table costs $+0.027$ with an interval from $-0.006$ to $+0.065$ ($p = 0.22$): the data are consistent with no cost and with a cost of six points, most of the $+0.046$ ($[+0.016, +0.081]$) that gold conditioning gives over the baseline on these examples. What can be stated is a bound: the zeroed additive model is no more than one point below the flat baseline (zeroed minus A: $+0.019$, $[-0.006, +0.044]$). A single-seed attention run that we had first read as showing the embedding could be removed without loss (Appendix~\ref{app:attention}) was underpowered.

D-token differs. Every perturbation lowers its accuracy with a bootstrap interval excluding zero, including the class mean ($+0.073$, $[+0.046, +0.100]$), though with five seeds only the cyclic wrong type also passes the seed-level test ($p = 0.001$). Its zeroed model is below the flat baseline by $-0.067$ ($[-0.089, -0.044]$, seed $p < 0.01$), whereas the zeroed D-hook is not: adaptation with a prepended token makes the model depend on it, adaptation with an additive vector does not, even though the vector is used. The prompt word behaves the same way (Table~\ref{tab:interv_dtext}; three seeds): a wrong word collapses D-text from 0.312 to 0.076 and 0.085 (gold minus wrong $+0.236$ and $+0.227$, seed $p = 0.017$ and $0.014$), and with no word it scores 0.205, below the baseline by $-0.049$ ($[-0.079, -0.018]$). Under a wrong type, the failure a deployed classifier produces, the additive model is the most brittle: 0.054 and 0.099 against D-token's 0.079 and 0.155 and D-text's 0.076 and 0.085. Conditioning acts almost entirely on clicks in every mechanism, and attention at the y-predicting token moves with accuracy under these interventions on one seed (Appendix~\ref{app:attention}).

\subsection{A positional-encoding failure that inverted a conclusion}
\label{sec:mrope}
Our first implementation of the prepended token replaced the slot's embedding by building \texttt{inputs\_embeds} and passing that instead of \texttt{input\_ids}. On \mix{taps\_and\_swipes} it scored $0.298 \pm 0.026$ \hitat{0.10} against the flat baseline's $0.390 \pm 0.010$ in the same runs, and the same slot with a frozen random embedding scored 0.290, which pointed at the injection path. Qwen2-VL derives its three-dimensional rotary positions from \texttt{input\_ids}; given \texttt{inputs\_embeds} alone, transformers falls back to one-dimensional positions replicated across the three axes, and the image tokens lose their two-dimensional layout (Appendix~\ref{app:mrope}; \citealp{hfissue35463}). A forward hook with \texttt{input\_ids} intact removed the deficit. We inferred the cause from the library source and the frozen-embedding control rather than from position ids in the failing runs, and record it because the failure is silent.

\section{Limitations}
\label{sec:limitations}
The study is small: one 2B backbone, LoRA adapters, validation slices of 200 to 250 steps from 35 to 50 episodes, one learning rate and adapter configuration for every variant, a stored-order split rather than the official episode split, and no action history. Per-seed variance is large, so the conclusions rest on paired statistics at two levels, neither clean: the episode bootstrap ignores that seeds share a data order, and the seed-level test has five or three units. The prepended token's failure at the shared learning rate is a schedule effect and the faster-trained token has three seeds; we did not train a frozen-slot control on the fixed path, and the class-mean intervention is matched to D-token in direction but not in norm. The clean-stream result bounds rather than excludes a small advantage, and how far the degenerate-class hazard generalizes is open: whether other corpora encode non-spatial actions the same damaging way, whether the effect scales, and whether the residual mean-distance advantage holds on larger slices. Those are the experiments we would run next. The end-to-end result is descriptive, the scaling study is uncorrected on moving slices, the attention analysis uses one seed and one layer, and Mind2Web is at floor. Code, split indices, prompts, and the per-example logs behind every table are available at \url{https://github.com/aadcha/action-conditioned-gui-agent}.

\section{Conclusion}
\label{sec:conclusion}
An auxiliary loss, an additive embedding, and a word in the prompt each help a small grounding model on a stream our serialization gave a degenerate class; none is established to help once that class is removed or on taps and swipes, and the prepended token we set out to validate helps only when its table is trained faster than the shared schedule, and then no better than the simpler mechanisms. Both learned embeddings are read at inference: zeroed, the additive model stays within a point of the baseline and the prepended one falls seven points below, and a wrong type collapses both, so the classifier is the risk; with an 84\% one the pipeline's margin over a flat model is not established. The practical lessons: audit what a serialization does to every class before supervising a coordinate on it, prefer the prompt word or the auxiliary loss over new parameters, perturb the conditioning to test whether it is read, and keep \texttt{input\_ids} on the path whenever conditioning touches the embeddings.

\bibliographystyle{plainnat}
\bibliography{references}

\clearpage
\appendix

\section{Training and evaluation details}
\label{app:details}
\paragraph{Data.} AITW screenshots in the \texttt{cjfcsjt/AITW\_General} mirror are stored as raw RGB byte buffers without an image header; we decode them by matching the buffer length against seven observed resolutions (for example $540\times1140$ and $412\times732$) with a portrait-aspect fallback. Mind2Web screenshots are downscaled so that no image exceeds one megapixel. Examples are drawn from the AITW training stream in stored order, filtered to the mix's action labels; the first $n$ eligible steps form the training slice and the next 250 (200 for the control mix) form validation. Consecutive steps belong to the same episodes, so a validation slice spans a few dozen episodes (35 to 50 across the sizes we use) and shares its boundary episode with training (two to ten steps, four at the headline size). Table~\ref{tab:scaling} lists the class mix of each validation slice.

\paragraph{Prompts.} Every Stage-2 prompt is the chat-template user turn containing the screenshot, the line \texttt{Goal: \{goal\}}, and the instruction \texttt{Predict the action coordinate.}; the assistant turn is the answer \texttt{(x, y)} with integers on the $0$--$999$ grid. C uses \texttt{Predict the action type and coordinate.} and is trained to answer with the action word before the coordinate, \texttt{click (x, y)}; D-text prefixes the user turn with \texttt{Action: \{word\}.}; D-token places \texttt{<|action\_slot|>} at the start of the text, and exactly one slot per sequence is asserted at every step.

\paragraph{Optimization.} AdamW with learning rate $2\times10^{-5}$ and weight decay 0.01, batch size 1, gradient clipping at 1.0, two epochs, a fixed per-epoch shuffle seeded by the run seed. The action-embedding table (D-hook, D-token) and the auxiliary head (B) share the base learning rate unless stated otherwise. Evaluation is greedy with at most 16 new tokens; the coordinate is parsed with a regular expression and an unparseable output counts as a miss at distance $\sqrt{2}$. Greedy evaluation is deterministic for a trained model; models in different tables that share a seed are separately trained copies (Table~\ref{tab:causal} and the intervention runs retrain the variant), and their numbers differ by training nondeterminism.

\paragraph{Compute.} Everything runs on single NVIDIA L4 GPUs. A Stage-2 run at 1,200 training examples takes about 35 minutes; the full study, including retracted and re-run cells, cost about 150 GPU-hours.

\paragraph{Statistics.} For each contrast we align the per-example distances of the two variants on every shared seed. The episode bootstrap resamples the episodes of the validation slice with replacement, keeps every step and every seed of a sampled episode together, recomputes the mean difference 10,000 times, and reports the percentile interval and the two-sided bootstrap $p$. The seed-level test is a paired $t$-test on the per-seed means. Clustering on examples instead of episodes would add one star in the paper (hard routing's headline gain, example-cluster interval $[+0.001, +0.069]$); the (seed, example)-unit bootstrap and its permutation counterpart give intervals about a third narrower still and are reported in the released logs. Every star in this paper was recomputed on the episode basis.

\section{Stage 1, Mind2Web, and per-class tables}
\label{app:tables}

\begin{table}[h]
\centering
\small
\caption{Stage~1 action-type macro-F1 (three seeds). On Mind2Web the instruction names the action, so the screenshot adds little; on AITW the instruction describes a goal and the screenshot is what separates the classes.}
\label{tab:stage1}
\begin{tabular}{lcc}
\toprule
Features & Mind2Web (2 classes) & AITW (5 classes) \\
\midrule
text only & $0.597 \pm 0.011$ & $0.141 \pm 0.032$ \\
text + screenshot & $0.605 \pm 0.016$ & $0.555 \pm 0.036$ \\
zeroed (sanity) & $0.469 \pm 0.000$ & $0.110 \pm 0.051$ \\
\midrule
TF-IDF + logistic regression & $0.622$ & $0.168$ \\
majority class & $0.472$ & $0.140$ \\
\bottomrule
\end{tabular}
\end{table}

\begin{table}[h]
\centering
\small
\caption{Multimodal-Mind2Web grounding control (1,000 training and 250 evaluation steps, three seeds, seed-level only). The target is the center of the gold element; hit@bbox is the fraction of predicted points inside the element. Both models are at floor.}
\label{tab:m2w}
\begin{tabular}{lcccc}
\toprule
Variant & \hitat{0.10} & \hitat{0.25} & hit@bbox & mean L2 $\downarrow$ \\
\midrule
A: flat & $0.365 \pm 0.031$ & $0.645 \pm 0.035$ & $0.095 \pm 0.018$ & $0.234 \pm 0.023$ \\
D-hook: additive & $0.355 \pm 0.032$ & $0.664 \pm 0.052$ & $0.104 \pm 0.034$ & $0.233 \pm 0.027$ \\
\bottomrule
\end{tabular}
\end{table}
\FloatBarrier

\section{Type events removed from training}
\label{app:notype}
Table~\ref{tab:notype} retrains all six variants on the headline training slice with the 35 type events removed, scored on the identical 250 validation examples, so the with-type and type-removed rows of each variant are paired example by example.
\begin{table}[h]
\centering
\small
\caption{Removing \emph{type} events from the training slice only, scored on the identical headline validation slice (3 to 5 seeds; each contrast uses the seeds its two conditions share, given per row). Deltas are paired episode-cluster bootstraps (every step and seed of a sampled episode kept together); seed-level $p$ in parentheses.}
\label{tab:notype}
\resizebox{\linewidth}{!}{\begin{tabular}{llccccc}
\toprule
Variant & Training stream & seeds & hit@0.10 & click hit@0.10 & scroll hit@0.10 & $\Delta$hit@0.10 vs.\ A (same stream) \\
\midrule
A: flat & with type & 5 & 0.229 $\pm$ 0.043 & 0.256 $\pm$ 0.062 & 0.304 $\pm$ 0.034 &  \\
A: flat & type removed & 5 & 0.296 $\pm$ 0.019 & 0.355 $\pm$ 0.031 & 0.304 $\pm$ 0.051 &  \\
B: aux.\ loss & with type & 5 & 0.293 $\pm$ 0.021 & 0.357 $\pm$ 0.026 & 0.278 $\pm$ 0.018 &  \\
B: aux.\ loss & type removed & 5 & 0.298 $\pm$ 0.038 & 0.368 $\pm$ 0.069 & 0.265 $\pm$ 0.058 & +0.002 [-0.014, +0.018] (0.926) \\
C: hard routing & with type & 5 & 0.264 $\pm$ 0.045 & 0.350 $\pm$ 0.061 & 0.148 $\pm$ 0.073 &  \\
C: hard routing & type removed & 3 & 0.269 $\pm$ 0.033 & 0.355 $\pm$ 0.039 & 0.159 $\pm$ 0.126 & -0.028 [-0.071, +0.013] (0.118) \\
D-hook: additive & with type & 5 & 0.282 $\pm$ 0.041 & 0.321 $\pm$ 0.064 & 0.357 $\pm$ 0.039 &  \\
D-hook: additive & type removed & 5 & 0.297 $\pm$ 0.039 & 0.341 $\pm$ 0.061 & 0.361 $\pm$ 0.029 & +0.001 [-0.020, +0.022] (0.973) \\
D-token: prepended & with type & 5 & 0.238 $\pm$ 0.049 & 0.269 $\pm$ 0.069 & 0.304 $\pm$ 0.090 &  \\
D-token: prepended & type removed & 3 & 0.273 $\pm$ 0.015 & 0.333 $\pm$ 0.028 & 0.261 $\pm$ 0.038 & -0.024 [-0.040, -0.008]$^{**}$ (0.402) \\
D-text: word in prompt & with type & 5 & 0.302 $\pm$ 0.028 & 0.348 $\pm$ 0.036 & 0.361 $\pm$ 0.036 &  \\
D-text: word in prompt & type removed & 3 & 0.283 $\pm$ 0.017 & 0.318 $\pm$ 0.015 & 0.370 $\pm$ 0.038 & -0.015 [-0.064, +0.036] (0.490) \\
\bottomrule
\end{tabular}}
\end{table}

\FloatBarrier

\section{Scaling study}
\label{app:scaling}
\begin{figure}[h]
  \centering
  \includegraphics[width=0.85\linewidth]{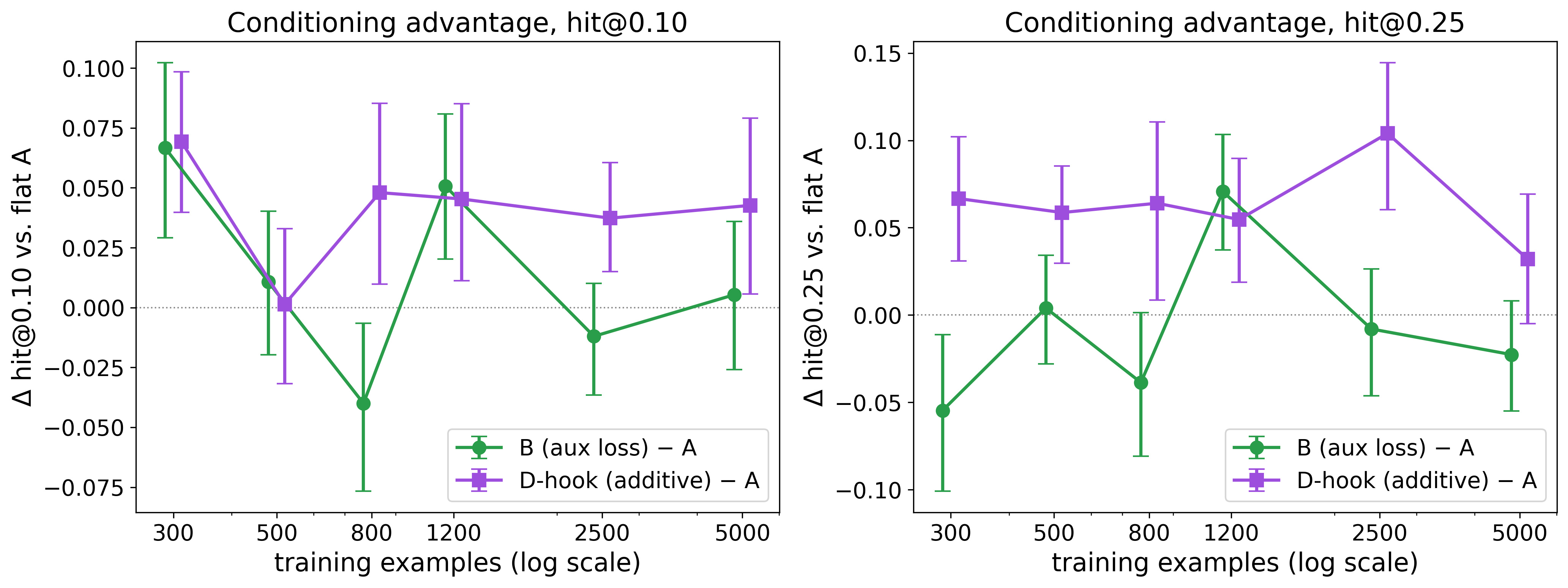}
  \caption{Conditioning advantage over the flat baseline as a function of training-set size on \mix{all\_with\_coords}, three seeds per cell, 95\% episode-clustered bootstrap intervals. Each size is scored on a different validation slice with a different class mix, so only within-size differences are meaningful, and 24 contrasts are shown without correction.}
  \label{fig:scaling}
\end{figure}

Figure~\ref{fig:scaling} and Table~\ref{tab:scaling} extend the comparison of A, B, and D-hook to $n \in \{300, 500, 800, 1200, 2500, 5000\}$ with three seeds per size. Each size is scored on a different validation slice whose scroll count ranges from 12 to 71, and B hurts scrolls while D-hook helps them (Table~\ref{tab:perclass}), so only within-size contrasts are read, and the 24 contrasts are uncorrected: D-hook $-$ A excludes zero at five of the six sizes (all but 500) on \hitat{0.10}, B $-$ A at 300 and 1,200 and is negative at 800, and with three seeds the seed-level tests pass only at 300 and 5,000 for D-hook and at 300 and 1,200 for B. An earlier single-seed curve had suggested that every conditioning gain shrinks with data, which three seeds do not support.
\begin{table}[h]
\centering
\small
\caption{Conditioning advantage vs.\ training-set size (AITW \texttt{all\_with\_coords}, 3 seeds per cell). Intervals are 95\% episode-cluster bootstraps (every step and seed of a sampled episode kept together); stars from the same bootstrap. Absolute values are not comparable across rows because the validation slice moves with $n$; its click/scroll/type counts are given per row. The study is descriptive: 24 contrasts, no multiplicity correction.}
\label{tab:scaling}
\resizebox{\linewidth}{!}{\begin{tabular}{rlccccc}
\toprule
$n_{\text{train}}$ & val mix (c/s/t) & A hit@0.10 & B $-$ A & D-hook $-$ A & B $-$ A (hit@0.25) & D-hook $-$ A (hit@0.25) \\
\midrule
300 & 170/33/47 & 0.139 $\pm$ 0.032 & +0.067 [+0.029, +0.102]$^{**}$ & +0.069 [+0.040, +0.098]$^{***}$ & -0.055 [-0.101, -0.011]$^{*}$ & +0.067 [+0.031, +0.102]$^{***}$ \\
500 & 198/12/40 & 0.315 $\pm$ 0.059 & +0.011 [-0.020, +0.040] & +0.001 [-0.032, +0.033] & +0.004 [-0.028, +0.034] & +0.059 [+0.030, +0.085]$^{***}$ \\
800 & 141/71/38 & 0.261 $\pm$ 0.020 & -0.040 [-0.077, -0.006]$^{*}$ & +0.048 [+0.010, +0.085]$^{*}$ & -0.039 [-0.081, +0.001] & +0.064 [+0.009, +0.111]$^{*}$ \\
1200 & 169/46/35 & 0.255 $\pm$ 0.026 & +0.051 [+0.020, +0.081]$^{***}$ & +0.045 [+0.011, +0.085]$^{**}$ & +0.071 [+0.037, +0.103]$^{***}$ & +0.055 [+0.019, +0.090]$^{**}$ \\
2500 & 166/53/31 & 0.244 $\pm$ 0.037 & -0.012 [-0.037, +0.010] & +0.037 [+0.015, +0.061]$^{**}$ & -0.008 [-0.046, +0.026] & +0.104 [+0.060, +0.144]$^{***}$ \\
5000 & 187/29/34 & 0.363 $\pm$ 0.009 & +0.005 [-0.026, +0.036] & +0.043 [+0.006, +0.079]$^{*}$ & -0.023 [-0.055, +0.008] & +0.032 [-0.005, +0.069] \\
\bottomrule
\end{tabular}}
\end{table}

\FloatBarrier

\section{D-token: learning-rate sweep and the earlier three-condition test}
\label{app:causal}
\begin{table}[h]
\centering
\small
\caption{D-token with the action-embedding table trained at 1$\times$, 10$\times$, and 100$\times$ the base learning rate on the headline slice. Displacement is the mean distance the rows of the classes present (click, type, scroll) moved from their $\mathcal{N}(0, 0.02^2)$ initialization, whose norm is 0.78; the 1$\times$ value comes from the type-removed runs, which share the code path and learning rate. $\Delta$ vs.\ A uses the shared seeds with an episode-cluster interval.}
\label{tab:dtoken_lr}
\resizebox{\linewidth}{!}{\begin{tabular}{lccccccc}
\toprule
Table LR & seeds & hit@0.10 & hit@0.25 & click hit@0.10 & $\Delta$hit@0.10 vs.\ A & seed $p$ & row displacement \\
\midrule
shared ($2\times10^{-5}$) & 5 & 0.238 $\pm$ 0.049 & 0.504 $\pm$ 0.055 & 0.269 $\pm$ 0.069 & +0.009 [-0.019, +0.036] & 0.465 & 0.016 \\
$2\times10^{-4}$ (10$\times$) & 3 & 0.295 $\pm$ 0.041 & 0.561 $\pm$ 0.045 & 0.329 $\pm$ 0.048 & +0.040 [+0.005, +0.076]$^{*}$ & 0.157 & 0.276 \\
$2\times10^{-3}$ (100$\times$) & 3 & 0.283 $\pm$ 0.032 & 0.544 $\pm$ 0.022 & 0.318 $\pm$ 0.030 & +0.028 [-0.008, +0.065] & 0.020 & 1.874 \\
\bottomrule
\end{tabular}}
\end{table}

Table~\ref{tab:dtoken_lr} is the sweep behind Section~\ref{sec:headline}: the same D-token architecture with only the action-embedding table's learning rate changed, three seeds per rate against the shared seeds of the flat baseline. The displacement column is measured on runs that log the table before and after training; the headline runs predate that logging, so the shared-rate value comes from the type-removed runs, which use the same rate and code path.
\begin{table}[h]
\centering
\small
\caption{D-token causal-use test: the same trained model evaluated with the gold action embedding, a wrong (cyclically permuted) one, and a zeroed one (5 seeds; paired episode-cluster bootstrap intervals).}
\label{tab:causal}
\resizebox{\linewidth}{!}{\begin{tabular}{lcccc}
\toprule
Conditioning & hit@0.05 & hit@0.10 & hit@0.25 & mean L2 $\downarrow$ \\
\midrule
gold & 0.140 $\pm$ 0.059 & 0.236 $\pm$ 0.058 & 0.502 $\pm$ 0.045 & 0.386 $\pm$ 0.020 \\
wrong & 0.014 $\pm$ 0.004 & 0.079 $\pm$ 0.013 & 0.200 $\pm$ 0.027 & 0.728 $\pm$ 0.081 \\
zero & 0.074 $\pm$ 0.055 & 0.167 $\pm$ 0.066 & 0.416 $\pm$ 0.103 & 0.489 $\pm$ 0.059 \\
\midrule
gold $-$ wrong & & +0.157 [+0.120, +0.192]$^{***}$ & +0.302 [+0.258, +0.342]$^{***}$ & -0.342 [-0.366, -0.314]$^{***}$ \\
gold $-$ zero & & +0.069 [+0.042, +0.096]$^{***}$ & +0.086 [+0.062, +0.109]$^{***}$ & -0.103 [-0.115, -0.091]$^{***}$ \\
\bottomrule
\end{tabular}}
\end{table}

Table~\ref{tab:causal} is the test that preceded Table~\ref{tab:interv}: separately trained D-token models (five seeds), decoded with the gold embedding, the cyclic wrong map, and the table zeroed. Its numbers differ from the D-token rows of Table~\ref{tab:interv} by training nondeterminism and agree in every conclusion. The class-mean control of Table~\ref{tab:interv} is matched to D-token in direction but not in norm: the mean of three near-orthogonal rows of norm 0.78 has norm about 0.45, so a fresh $\mathcal{N}(0, 0.02^2)$ draw would be the matched control, and we did not run it.
\FloatBarrier

\section{Interventions on the prompt word}
\label{app:interv_dtext}
\IfFileExists{tables/interv_dtext.tex}{\begin{table}[h]
\centering
\small
\caption{Interventions on D-text: the same trained model decoded with the gold action word, a wrong word (cyclic; click$\to$type), a wrong word swapping only clicks and scrolls, and no action word (3 seeds $\times$ 250 examples; $\Delta$ = gold $-$ condition, 95\% episode-cluster CI, seed-level paired $p$; flat A on the same examples is the reference row).}
\label{tab:interv_dtext}
\resizebox{\linewidth}{!}{\begin{tabular}{lccccc}
\toprule
Conditioning & hit@0.10 & click hit@0.10 & scroll hit@0.10 & $\Delta$ gold $-$ cond.\ (hit@0.10) & seed $p$ \\
\midrule
gold word & 0.312 $\pm$ 0.045 & 0.357 $\pm$ 0.056 & 0.384 $\pm$ 0.045 &  &  \\
wrong word (cyclic) & 0.076 $\pm$ 0.011 & 0.028 $\pm$ 0.003 & 0.312 $\pm$ 0.050 & +0.236 [+0.183, +0.289]$^{***}$ & 0.017 \\
wrong word (click$\leftrightarrow$scroll) & 0.085 $\pm$ 0.006 & 0.041 $\pm$ 0.006 & 0.312 $\pm$ 0.050 & +0.227 [+0.174, +0.281]$^{***}$ & 0.014 \\
no action word & 0.205 $\pm$ 0.067 & 0.217 $\pm$ 0.102 & 0.319 $\pm$ 0.055 & +0.107 [+0.063, +0.152]$^{***}$ & 0.241 \\
flat A, same examples & 0.229 $\pm$ 0.043 & 0.256 & 0.304 & +0.057 [+0.018, +0.099]$^{**}$ & 0.083 \\
\bottomrule
\end{tabular}}
\end{table}
}{}
Table~\ref{tab:interv_dtext} applies the protocol of Section~\ref{sec:interventions} to D-text, whose conditioning is a word in the prompt rather than a learned vector: the same trained model is decoded with the gold word, a wrong word under the two maps, and no word at all.
\FloatBarrier

\section{The \texttt{inputs\_embeds} failure in detail}
\label{app:mrope}
Qwen2-VL locates the image tokens in \texttt{input\_ids} to assign each visual token a temporal, height, and width position for its three-dimensional rotary embedding. When a caller supplies \texttt{inputs\_embeds} and no \texttt{input\_ids}, the model cannot perform that lookup; in the transformers releases our runs used (pinned at 4.45 or later; the current 5.9 release behaves the same) the positions become a one-dimensional sequence index replicated across the three axes, so every image token is positioned as text and the two-dimensional layout the grounding task depends on is lost. A caller on that path must also merge the visual features into the embeddings itself, a second opportunity for silent error. We inferred the cause from the code and from the frozen-embedding control described in Section~\ref{sec:mrope} rather than by dumping position ids at the time. The same dependence on \texttt{input\_ids} was reported as a generation crash in December 2024~\citep{hfissue35463}; the silent degradation when images are present is the case a fine-tuning practitioner meets. In the set of runs where the failure appeared, the fixed additive variant scored $0.392 \pm 0.043$ \hitat{0.10} on \mix{taps\_and\_swipes} against the flat baseline's $0.390 \pm 0.010$; the re-run of that control with per-example logging (Table~\ref{tab:control}) gives $0.363 \pm 0.021$ against $0.382 \pm 0.015$.

\section{Where the model looks}
\label{app:attention}

\begin{figure}[h]
  \centering
  \includegraphics[width=0.9\linewidth]{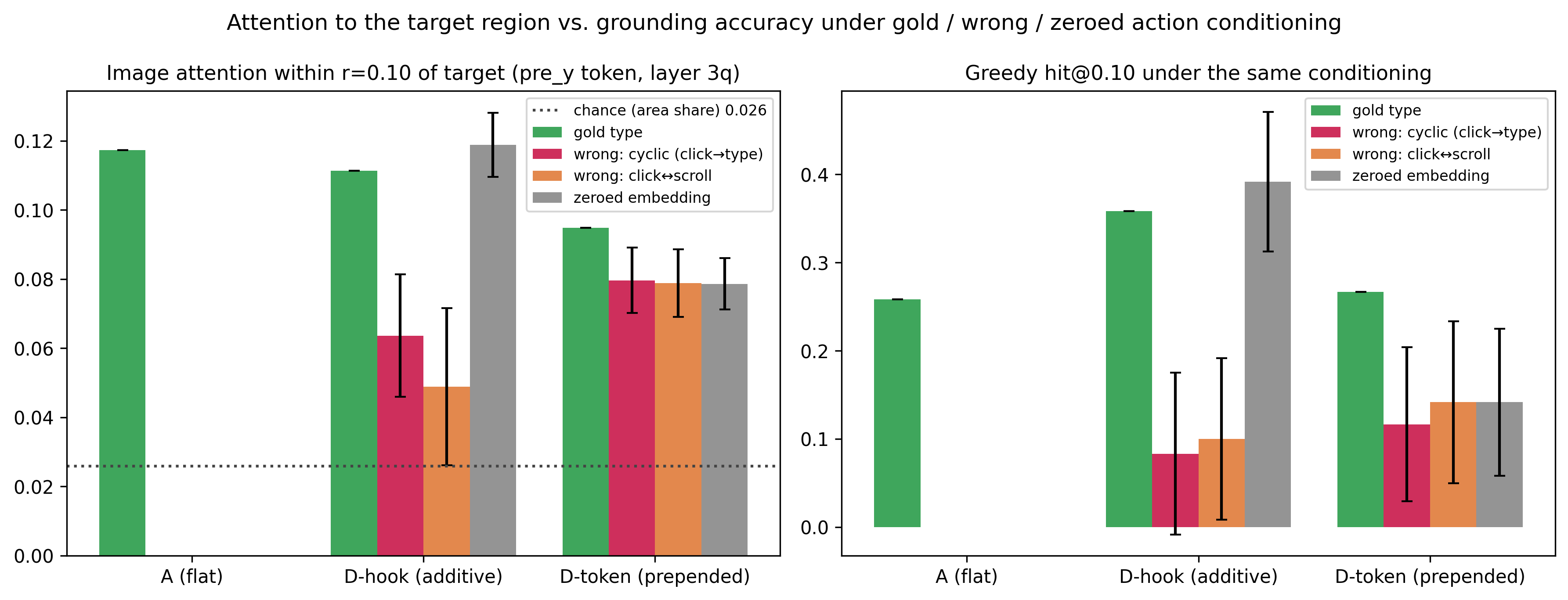}
  \caption{Same trained model, four conditionings, 120 validation screens, one seed per variant. Left: share of image attention within $0.10$ of the target from the token that predicts the y coordinate, with the gold answer teacher-forced (3/4-depth layer, head-averaged); the dotted line is the share of image tokens within that radius. Right: greedy \hitat{0.10} under the same conditioning. Error bars are 95\% paired-bootstrap intervals of the difference from gold over the 120 screens.}
  \label{fig:attn}
\end{figure}

We measured where the models look, with the caveats that the measure is head-averaged attention at one layer, that attention is not information flow, and that teacher-forcing the gold answer places the gold x coordinate in the prefix of the token that predicts y. Under gold conditioning both embedding models put 10 to 11\% of that token's image attention within $0.10$ of the target, against a chance share of 2.6\%; the last prompt token, which our earlier visualizations used, puts about 4\% there under every conditioning, so most of the localization we can see happens after the x coordinate is in the prefix. Teacher-forcing the model's own answer instead of the gold one, so that the y-predicting token sees the predicted x, lowers the target share only slightly (D-hook 0.092, D-token 0.085), so the localization is not an artifact of the supplied x. The flat baseline trained on the same seed localizes just as much (0.117 at the y-predicting token, 0.086 free-running) while grounding far less accurately on the same screens (0.258 vs.\ 0.358 \hitat{0.10}), so conditioning does not add localization; the attention share is a property of the fine-tuned model, not of the type signal. \citet{prefill} report that grounding is largely decided during prefill; our probe positions are answer tokens under teacher forcing, so the two measurements are not directly comparable. The interventions move attention the same way they move accuracy on this seed (Figure~\ref{fig:attn}, Table~\ref{tab:attn_positions}): a wrong type roughly halves the target share for D-hook, and zeroing costs D-token both attention and accuracy. On this one seed zeroing left D-hook's attention and accuracy unchanged; we had first read that as showing the embedding could be removed without loss, and the five-seed study in Section~\ref{sec:interventions} bounds the cost instead. Table~\ref{tab:attn_positions} gives the target-attention share for the last prompt token (the one that predicts the opening parenthesis), the token that predicts the first x digit, and the token that predicts the first y digit.

\begin{table}[h]
\centering
\small
\caption{Share of image attention within $0.10$ of the target (chance $0.026$) at three probe positions, seed 42, 120 screens, 3/4-depth layer, gold answer teacher-forced.}
\label{tab:attn_positions}
\begin{tabular}{llcccc}
\toprule
Model & Position & gold & wrong (cyclic) & wrong (click$\leftrightarrow$scroll) & zero \\
\midrule
A (flat) & last prompt token & 0.038 & & & \\
A (flat) & pre-y token & 0.117 & & & \\
D-hook & last prompt token & 0.040 & 0.036 & 0.035 & 0.036 \\
D-hook & pre-x token & 0.039 & 0.036 & 0.037 & 0.039 \\
D-hook & pre-y token & 0.111 & 0.064 & 0.049 & 0.119 \\
D-token & last prompt token & 0.038 & 0.034 & 0.034 & 0.035 \\
D-token & pre-x token & 0.038 & 0.034 & 0.033 & 0.035 \\
D-token & pre-y token & 0.095 & 0.080 & 0.079 & 0.079 \\
\bottomrule
\end{tabular}
\end{table}
\FloatBarrier

\section{Qualitative examples}
\label{app:qual}
\begin{figure}[h]
  \centering
  \includegraphics[width=\linewidth]{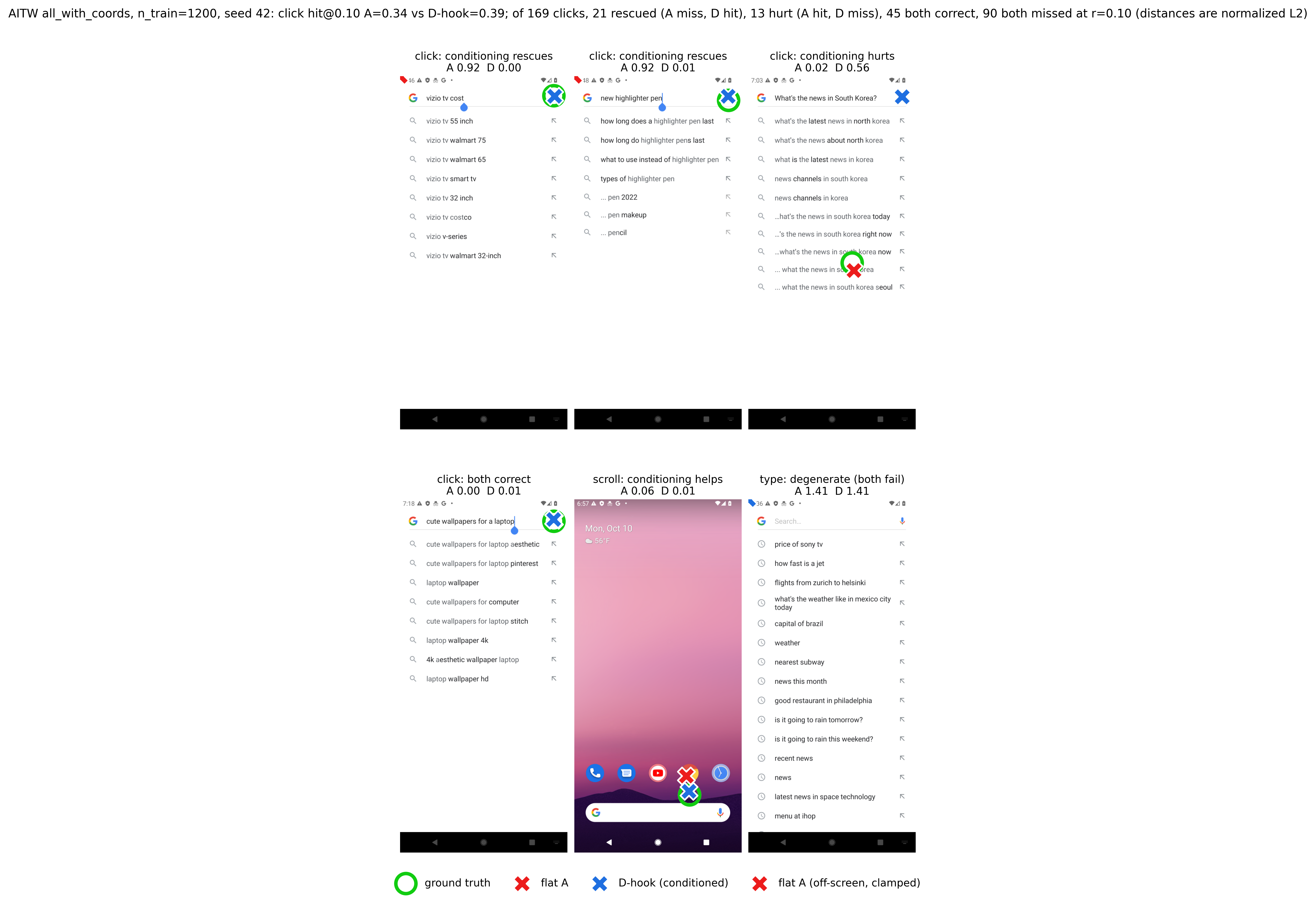}
  \caption{Predictions of the flat baseline (red) and D-hook (blue) against the target (green) on held-out AITW screens at the headline setting, one seed. The panels are a fixed spread of outcomes rather than a selection of wins: two rescues, a case where conditioning hurts, a case where both are right, a scroll, and a degenerate \emph{type} example on which both models miss at distance $\sqrt{2}$. The title reports the base rates on this seed at $r = 0.10$: of 169 clicks, 21 are rescued by conditioning, 13 are hurt, 45 are correct under both models, and 90 are missed by both. In the rescue panels the baseline's prediction is off-screen and drawn clamped at the corner.}
  \label{fig:qualitative}
\end{figure}

Figure~\ref{fig:qualitative} shows six held-out screens with the predictions of the flat baseline and of D-hook.

\end{document}